\documentclass[letterpaper, 10 pt, conference]{ieeeconf}  

\IEEEoverridecommandlockouts                              

\usepackage{cite}
\usepackage{amsmath,amssymb,amsfonts}
\usepackage{algorithmic}
\usepackage{graphicx}
\usepackage{textcomp}
\usepackage{booktabs} 
\usepackage{siunitx} 
\usepackage{svg}
\usepackage{bm}
\usepackage{array} 
\usepackage{xcolor} 

\title{\LARGE \bf

Imitation of human motion achieves natural head movements for humanoid robots in an active-speaker detection task}

\author{
Bosong Ding$^{1}$, Murat Kirtay$^{1}$, Giacomo Spigler$^{1}$
\thanks{$^{1}$ Department of Cognitive Science and Artificial Intelligence, Tilburg University, Tilburg, Netherlands. Corresponding author: {\tt\small \{b.ding\_3\}@tilburguniversity.edu}}%
}

\begin{document}

\maketitle
\thispagestyle{empty}
\pagestyle{empty}

\begin{abstract}

Head movements are crucial for social human-human interaction. They can transmit important cues (e.g., joint attention, speaker detection) that cannot be achieved with verbal interaction alone. This advantage also holds for human-robot interaction. Even though modeling human motions through generative AI models has become an active research area within robotics in recent years, the use of these methods for producing head movements in human-robot interaction remains underexplored. In this work, we employed a generative AI pipeline to produce human-like head movements for a Nao humanoid robot. In addition, we tested the system on a real-time active-speaker tracking task in a group conversation setting. Overall, the results show that the Nao robot successfully imitates human head movements in a natural manner while actively tracking the speakers during the conversation. Code and data from this study are available at {\color{cyan}https://github.com/dingdingding60/Humanoids2024HRI}.

\end{abstract}


\section{Introduction}

Head movements are important to generate nonverbal cues that enhance human-human interactions. For example, a student and teacher can construct joint attention by using head movements or a child can attend a speaker in group conversation to maintain a social interaction. Given the significance of head movement in group interaction, studies on human-robot interaction have also attempted to model natural head movements in interactive robots \cite{barot2023audio}. However, work on human-like motion generation for robots, particularly in real-time motion generation settings, remains less-explored.

Although modeling human (or animal) motion has been extensively explored in generative artificial intelligence, studies have so far been mostly limited to applications that involve locomotion, manipulation, and robot-robot interaction (e.g., playing soccer) \cite{physical_skills_from_videos}, \cite{npmp_humanoid}, \cite{laicago}, \cite{npmp_humanoid_and_quadruped}, \cite{liu2022motor}, \cite{aloha2}. Here, we adopt a generative modeling approach to imitate head movement of humans by using a Nao humanoid robot in a human-robot interaction (HRI) scenario. To be concrete, we formulate the following research question: \emph{How can we produce human-like head movements via generative modeling on a humanoid robot tasked to detect and track active speakers in an HRI setting?}

To answer this question, we designed HRI experiments in which a Nao humanoid robot participates group interaction to recognize active speakers. In this setting, the Nao robot generates head movement trajectories through a variational autoencoder (VAE), to reproduce motion similar to human demonstrations. The robot then passively participates in conversation with human partners by paying attention to the active speaker. We note that like most of the robots (e.g., Pepper) used in HRI studies, our robot can only move its head along the in yaw and pitch directions. Despite this limitation, we found that it is possible to match the corresponding human head movements, achieving natural-looking motion during the task. Moreover, we significantly improved the inference time for detecting the active speaker compared to state-of-the-art results, achieving an 85\% reduction in inference time—from 1.3 seconds to 0.2 seconds to process 1 second of video at 30 fps \cite{liao2023light}.


The novel contributions of this work are as follows. First, we developed an human-motion modeling pipeline on a Nao humanoid robot to generate natural head movement while detecting active speakers in a group conversation setting. We note that our proposed pipeline can also be deployed in different humanoid robots, such as the iCub or Pepper, to mention a few. Second, we present an extensive analysis of a proof-of-concept case study on an active-speaker interaction task.  Lastly, we provide a new dataset of human head-gaze motion together with the trained models and benchmarked results.

\section{Related Work} \label{related_work}

Although generative modeling of human motion is an active research area in robotics \cite{npmp_humanoid, npmp_humanoid_and_quadruped, liu2022motor}, its use for imitating head movements remains limited. For example, Grassi et al. \cite{grassi2023robot} developed predefined control policies for the Pepper robot in a group conversation setting. The authors showed that the balanced attention provided by the Pepper robot enhances conversational dynamics and reduces the likelihood of subgroup formation among the participants. Similarly, Barot et al. employed a REEM-C humanoid robot to interact with multiple participants to detect active speakers via their head movements using multimodal data. In addition to quantitative results, the authors provided a human subject study with 5 participants to assess interaction along three dimensions: naturalness, accuracy, and responsiveness \cite{barot2023audio}.

\begin{figure*}[ht!]
  \centering
  \includegraphics[width=\textwidth]{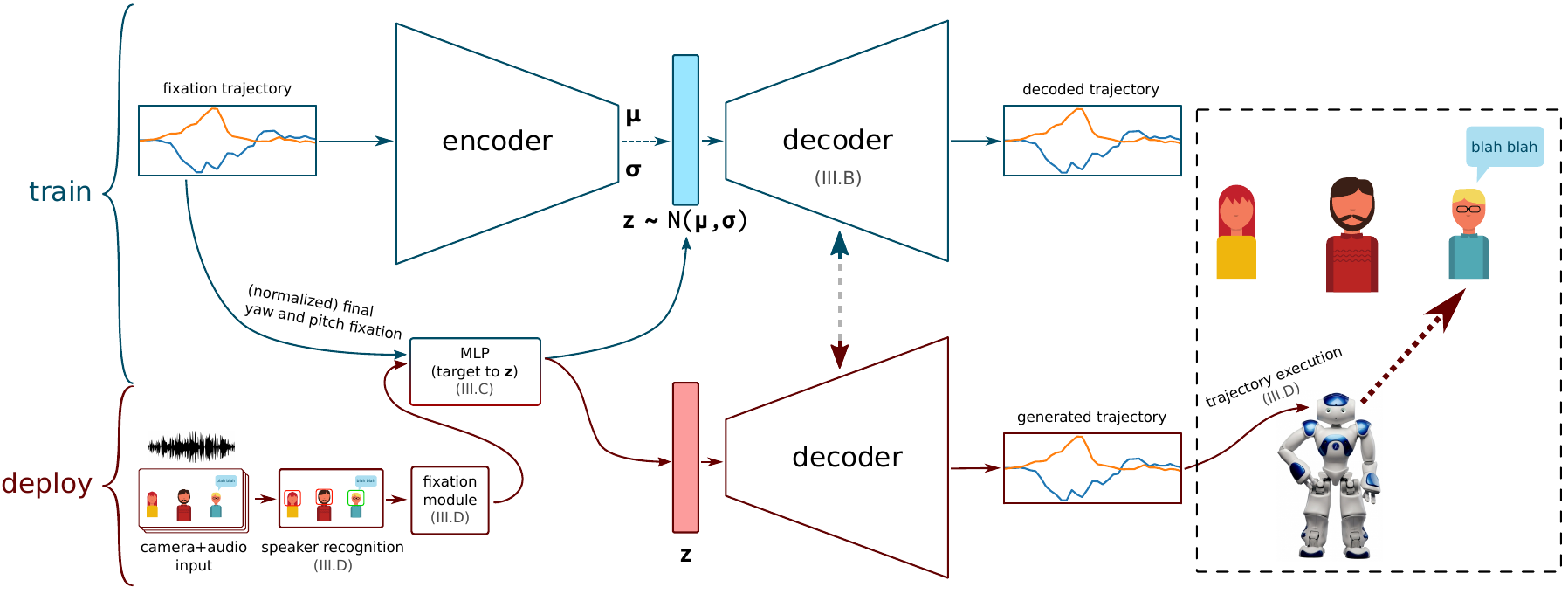}
  \caption{ Schematics of the proposed method for generating head-gaze movements from human motion data (methods \ref{sec:datacollection}), along with the example application to an active-speaker gazing task. The models are obtained by first training a variational autoencoder to model human motion data (methods \ref{sec:trajmodelling}). Next, a multilayer perceptron (MLP) is trained to map end fixation points of the training trajectories into the corresponding latent vectors $\mathbf{z}$ learned by the encoder (methods \ref{sec:traj_generation}). The system can then be used by first converting desired fixation points into latent vectors, which are subsequently transformed into motion trajectories by the VAE decoder. In the active-speaker fixation task (methods \ref{ASDsection}), audiovisual input from a camera mounted on the head of a Nao robot is inputted to the Light-ASD \cite{liao2023light} active-speaker recognition module to retrieve bounding boxes of faces together with a confidence score for each possible speaker. Fixations are then decided by sampling a softmax distribution of the confidence scores.}
  \label{fig:schematics}
\end{figure*}

Within the context of Active-Speaker Detection (ASD), previous studies have mostly focused on offline datasets as benchmarks, such as the Columbia ASD dataset \cite{chakravarty2016cross} and the AVA Active Speaker dataset \cite{roth2020ava}, rather than real-time use on social robots. For example, Liao et al. \cite{liao2023light} implemented an end-to-end deep learning architecture that outperforms different models, as evaluated using these two datasets.  Jung et al. \cite{jung2024talknce} proposed an active speaker detection model, TalkNCE, using contrastive learning with audiovisual data that achieves state-of-the-art results on the AVA-ActiveSpeaker dataset.  Alcazar et al. \cite{alcazar2022end} introduced an end-to-end network incorporating interleaved graph neural network blocks to aggregate spatio-temporal context using the AVA-ActiveSpeaker dataset. The above studies on active speaker detection utilize existing datasets. However, only a few studies were aimed at real-time active-speaker detection and they focused on hardware or voice localization \cite{gurvich2024real}, \cite{czarnecki2022we}.

Our work differs from previous work in the following way. First, unlike the head movement studies introduced above, we follow a generative modeling approach to produce \emph{natural} head movement for a Nao robot in a human-robot interaction setting. Second, our active-speaker detection pipeline was deployed on a Nao robot to achieve \emph{real-time} active speaker detection in group conversation experiments. Lastly, we conducted experiments in a realistic real-world setting where environmental noises and hardware constraints need to be considered and can affect the system's performance.

\section{Methods} \label{methods}

Modeling of human motion is achieved by first training a variational autoencoder (VAE) to learn the distribution of motion trajectories via unsupervised learning. Next, a multilayer perceptron (MLP) is trained to map end fixation points of the training trajectories into corresponding latent vectors learned by the VAE encoder. Evaluation of the generated motions is performed in an active-speaker gazing task, where target fixations are determined using a pre-trained active-speaker recognition model (Light-ASD \cite{liao2023light}). The overall system integration is shown in Fig. \ref{fig:schematics}.

\subsection{Head motion data collection}
\label{sec:datacollection}

We designed an experimental protocol to collect head movement data covering a large range of motion. A single participant wore a GoPro Hero 11 sports camera mounted on a head strap and positioned close to the eyes. During the experiment, data were collected from both the camera (not used in this study) in a linear lens mode at a resolution of 2.7K with a 4:3 aspect ratio, and from the inertial measurement unit (IMU) built into the camera. Data from both the camera and the IMU were sampled at 30 frames per second.


\begin{figure}[ht!]
  \centering
  \includegraphics[width=0.35\textwidth]{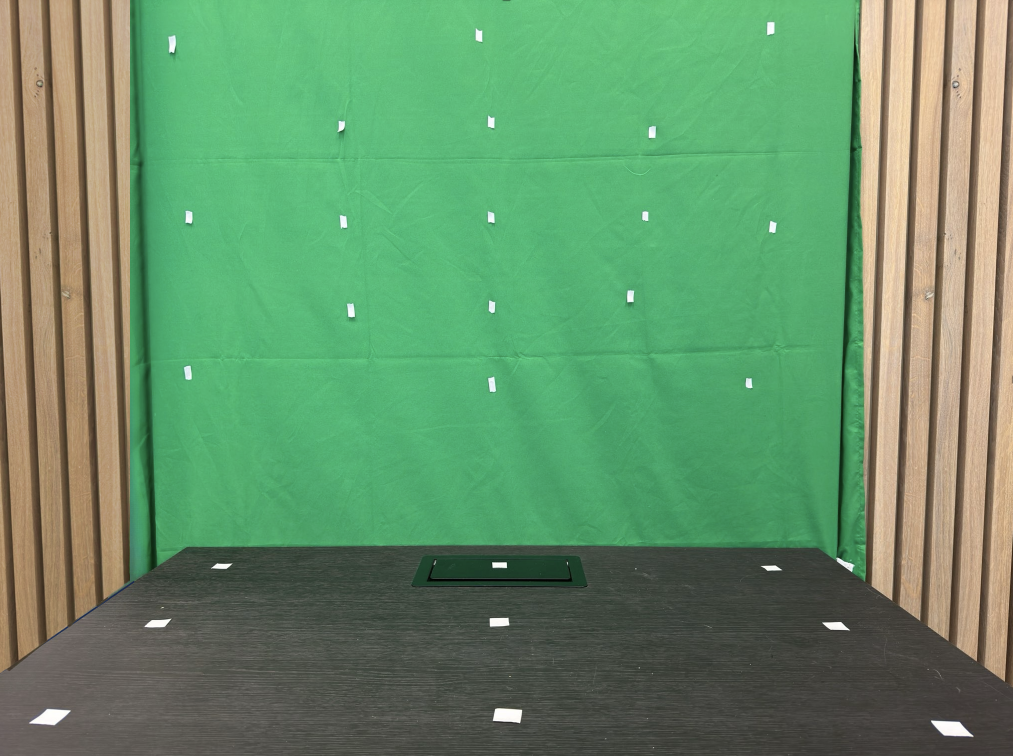}
  \caption{ Overview of the setup used to collect head movement trajectories (Section \ref{sec:datacollection}). A single participant is asked to produce head-gaze movements between pairs of points organized as three uniform $3\times 3$ grids of points. }
  \label{setup}
\end{figure}

The setup consisted of a table next to a wall, on top of whose were positioned three sets of evenly distributed 3x3 reference points to cover the visual field of the test subject. The setup is shown in Figure \ref{setup}. The first set covers a standard range directly in front of the subject, the second set spans a smaller range also in front, and the third set is positioned horizontally on the table. These arrangements of reference sets are designed to cover the majority of typical head movements. The distances between the reference points were determined based on the maximum rotation of the human head's yaw and pitch movements, as documented by Gilman et al. \cite{gilman1979measurement}. The smaller frontal range spans a range approximately half of the main one to imitate small and frequent head movement that often happens during human-robot interaction.

To ensure comprehensive spatial coverage, the participant was asked to systematically move through every possible combination of starting and ending points across each 3x3 grid, marking the beginning and end of each motion by pressing a button. To prevent any bias related to the relative positioning of the head and camera, all data is collected in a single session lasting approximately 30 minutes. During data collection, the participant was instructed to minimize the gaze movement as the Nao robot has no eye movement abilities. However, this setting is not necessarily different from human head motion during group conversation, since head movement is used by humans to optimize listening to the active speaker \cite{lu2022sound}. 
\begin{figure}[ht!]
  \centering
  \includegraphics[width=0.4\textwidth]{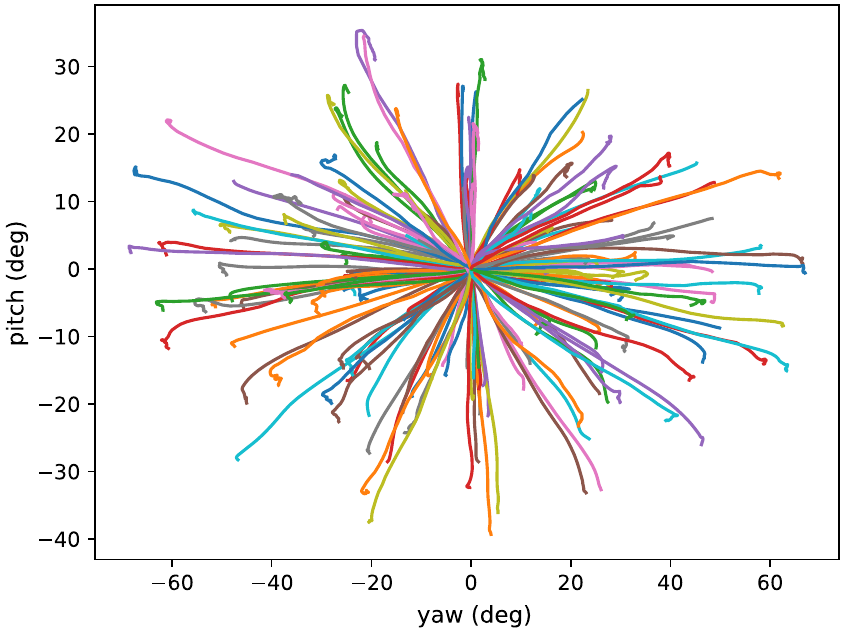}
  \caption{ Full set of the $174$ human head-motion trajectories collected. Trajectories consist of the yaw and pitch of the head during movement, relative to the initial pose (i.e., $\boldsymbol{\tau}(t) = \left(\mathrm{yaw}(t), \mathrm{pitch}(t) \right)$, $\boldsymbol{\tau}(0) = (0,0)$.) }
  \label{traj_all}
\end{figure}

Since the camera gyroscope records the absolute yaw and pitch angles of the head, we obtained a dataset of general head motions by subtracting the initial pose from each recorded trajectory. Figure \ref{traj_all} shows the full set of recorded trajectories.The dataset collected in this study is made available at our public repository\footnote{https://github.com/dingdingding60/Humanoids2024HRI}.

\subsection{Head trajectory modelling}
\label{sec:trajmodelling}

We took an approach similar to previous work on motion-capture trajectory modeling \cite{physical_skills_from_videos, npmp_humanoid, laicago}, and chose to model human head-movement trajectories using a variational autoencoder (VAE) \cite{vae}. Full architecture and hyperparameters are included in Appendix \ref{appendix:suppl_materials}.

We extracted individual trajectories from each trial in the training set by automatically identifying their beginning and end. The start of a fixation trajectory is detected by finding the first timestep where the magnitude of change $\| \Delta \boldsymbol{\tau} \|$ is above a threshold of $0.1$, then subtracting a fixed offset of 4 frames.  We used a sliding window of 20 frames to detect the end of each trajectory to eliminate small noise at the end of fixations. A fixation is marked to have ended once all the magnitudes within this window are less than 0.0175. Trajectories are then padded with trailing zeros until a fixed length of 60 frames (i.e., 2 seconds).

Trajectories $\boldsymbol{\tau}(t) = \left(\mathrm{yaw}(t), \mathrm{pitch}(t) \right)$ are represented as yaw and pitch angle of the head at each frame, relative to the initial pose (i.e., $\boldsymbol{\tau}(0) = (0,0)$ for every clip). We also approximate angular velocities at each frame by finite differences over the trajectory ($\Delta \boldsymbol{\tau}(t) = \boldsymbol{\tau}(t) - \boldsymbol{\tau}(t-1)$, where we assume $\Delta \boldsymbol{\tau}(0)=(0,0)$). An example of the trajectory of angular velocities $\Delta \boldsymbol{\tau}$ is shown in Figure \ref{fig:schematics}, as `fixation trajectory', and in the Appendix as Figure \ref{traj_vae_reconstruction_vel}.

A variational autoencoder is composed of two neural networks, an encoder $\mathbf{z_\mu}$, $\mathbf{z_\sigma} = \mathrm{enc}(\Delta \boldsymbol{\tau})$ that takes a trajectory of angular velocities as input and outputs two vectors, that parameterize a multivariate Normal distribution $\mathcal{N}(\mathbf{z_\mu}, \text{diag}(\mathbf{z_\sigma}))$, and a decoder $\mathrm{dec}(\mathbf{z})$ that takes a latent vector $\mathbf{z}$ sampled from the distribution proposed by the encoder, and outputs a reconstructed trajectory (as per-frame angular velocities) $\boldsymbol{\overline{\tau}}$ that is as similar as possible to the original trajectory.

We design a loss function that aims at reconstructing trajectories by matching both absolute $(\mathrm{yaw}, \mathrm{pitch})$ coordinates of the trajectory at each frame and the instantaneous angular velocities. The loss function used is

\begin{align*}
\scriptstyle
\hspace{-0.15in}
    \mathcal{L}_\mathrm{VAE} &= \underbrace{ \frac{1}{N} \sum_t \| \Delta\boldsymbol{\tau}(t) - \Delta\boldsymbol{\overline{\tau}}(t) \|^2 }_{\mathrm{velocity\, loss}} + \lambda_\mathrm{pos} \underbrace{ \frac{1}{N} \sum_t \| \boldsymbol{\tau}(t) - \boldsymbol{\overline{\tau}}(t) \|^2  }_{\mathrm{position\, loss}} \\
            &+ \lambda_\mathrm{KL} \, \mathrm{KL}\left( \mathcal{N}(\mathbf{z_\mu}, \text{diag}(\mathbf{z_\sigma})) \| \mathcal{N}(\mathbf{0}, I)  \right)
\end{align*}

where $\boldsymbol{\overline{\tau}} = \mathrm{dec}(\mathrm{enc}(\boldsymbol{\tau}))$ is a reconstructed trajectory, $N=60$ is the length of each trajectory in frames, $\lambda_\mathrm{pos}=5 \cdot 10^{-4}$ is a trade-off term between the MSE loss of absolute angles and the MSE loss of the per-frame angular velocities, and $\lambda_\mathrm{KL}=5 \cdot 10^{-3}$ is the relative strength of the VAE regularization term versus the reconstruction losses. We use a latent space of size 10, $\mathbf{z} \in \mathbb{R}^{10}$.


\subsection{Generation of target trajectories}
\label{sec:traj_generation}

The VAE is trained via unsupervised learning to match the distribution of human trajectories. To be useful in practical applications, however, we need a way to generate trajectories with desired properties. We are in particular interested in generating trajectories whose end point is at or close to a given target location. For example, if we wished to produce a fixation to a target $(\mathrm{yaw}, \mathrm{pitch}) = (-60^{\circ}, 20^{\circ})$, we would like to generate a trajectory similar to the pink one in the top-left corner of Figure \ref{traj_all}. Note that all trajectories are relative to the current fixation, given by an initial yaw and pitch angle.

We generate trajectories to given targets as follows. We process the dataset to calculate the end $(\mathrm{yaw}, \mathrm{pitch})$ fixation point for each clip of human trajectories. We then scale the fixation points by constant factors to approximately normalize the coordinates within $[-1, 1]$ (we use factors $\alpha_\mathrm{yaw}=\frac{1}{35}$, and $\alpha_\mathrm{pitch}=\frac{1}{25}$). We then train a multilayer perceptron (MLP) to map the normalized fixation targets into latent vectors $\mathbf{z}$ obtained by inputting the corresponding trajectories into the trained VAE \emph{encoder}. We provide Full architecture and hyperparameters in Appendix \ref{appendix:suppl_materials}.

At test time, trajectories can be generated by appropriately scaling target fixation points, inputting them into the MLP to obtain latent vectors that characterize the required motions, and then using the VAE \emph{decoder} to produce the desired trajectories.


\subsection{Case study: active-speaker gazing task}
\label{ASDsection}

We design a human-robot interaction scenario that involves the generation of repeated head movements to investigate the quality and naturality of the movements provided by our method, compared to a baseline where yaw/pitch head movements were driven by the default motion controller of the Nao robot. Due to the limitations of the narrow viewing angle (approximately $60^{\circ}$ horizontally) of the built-in camera and the need for wider visibility to effectively interact with multiple participants, we attached a Logitech webcam  ( $90^{\circ}$ field of view horizontally) to the top of Nao's head with a 3D-printed head mount based on Dhionis Sako's\footnote{https://github.com/costashatz/nao\_dcm} project.

We chose an Active Speaker Detection (ASD) task that involves identifying the active speaker in a group conversation with multiple potential speakers, in which the robot is a passive participant. The objective of ASD is to determine a function $f$ of an audio $A(t)$ and visual $V(t)$ data stream

\begin{equation}
    \mathbf{S}(t) = f(A(t), V(t))
\end{equation}

such that $\mathbf{S}(t) \in \{0,1\}^N$ is a vector whose components at each timestep $t$ are $S_i(t) = 1$ if speaker $i \in \{1,\dots, N\}$ is active and $0$ if it is not active.

Here, we use the pre-trained Light-ASD model from Liao et al.\cite{liao2023light}\footnote{https://github.com/Junhua-Liao/Light-ASD} to implement the function $\mathbf{S}(t)$. First, we detect bounding boxes for all faces seen by the Nao's camera in each frame. The detected faces are resized and organized into candidate speaker tracks using temporal Intersection Over Union (IOU) scores. For each candidate speaker, the sequence of face images and corresponding audio (shared across all speaker tracks) are processed independently through audio and visual encoders. The resulting audio and visual features are then concatenated. The combined feature vectors are then processed sequentially through a bidirectional Gated Recurrent Unit (GRU), followed by a multilayer perceptron.

Although Light-ASD is significantly faster than other state-of-the-art ASD models, we found it to still be too slow for use in real-time applications. This is likely due to the model having been developed primarily for offline processing of video files. As part of this work, we optimized the model by improving the preprocessing code to run in-memory and replacing the original face detection model with the faster MediaPipe \cite{mediapipe}. The combined modifications resulted in an 85\% reduction in overall inference time, decreasing the time required to process a 1s (at 30fps) video clip from 1.3 seconds to approximately 0.2 seconds. Furthermore, we designed an asynchronous system to record and process the robot's inputs in parallel, achieving close to 5 inferences per second. During interactions with humans, camera frames and audio chunks are stored in a First-In First-Out (FIFO) buffer that holds the most recent 1s of data. Every 0.5 seconds, the current buffer is asynchronously processed by the ASD model.

We finally implement a simple heuristic to select fixation targets as follows. Each time a new result is processed by the ASD module, we calculate a probability distribution by applying a softmax operator on the vector of scores assigned by the ASD module to each face in the last frame of the buffer, with a weighting factor $\beta=2$. Then, a new fixation is chosen as the center of the bounding box of the face sampled from the distribution. A motion trajectory is generated and executed only if the previous movement has already finished.

\subsection{ Preliminary experiment on human preferences }

We further test the subjective perception of the quality and naturalness of the generated motions by designing a preliminary experiment with, $n=9$, human participants. For the experiment, we collected $3$ videos centered on the Nao's head for both our method and a baseline motion controller. We then asked the participants to select 3 out of 6 videos with the following instructions: ``Choose 3 out of the 6 videos that in your opinion show the most natural and engaging movement of the Nao robot. Please note that head movements (i.e., target fixation points) were chosen using the \textbf{same} algorithm in all videos.''


\begin{figure*}[ht!]
  \centering
  \includegraphics[width=\textwidth]{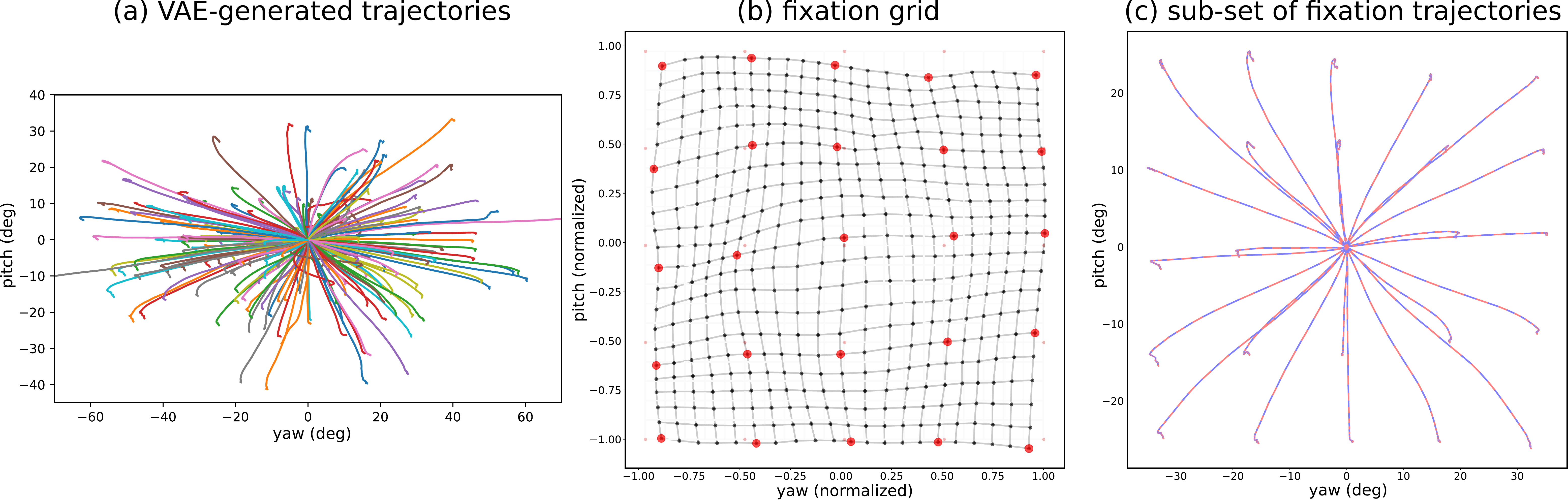}
  \caption{ Analysis of the capabilities of the proposed method to synthesize realistic head-gaze movements. (a) We generate and plot $200$ random trajectories by sampling latent vectors from the VAE prior distribution $\mathbf{z} \sim \mathcal{N}(\mathbf{0}, I)$, and compare them to the human motion data from Figure \ref{traj_all}. (b) We evaluate the capacity of the system to generate desired head-motion trajectories to look at target points. To do so, we select a set of $21 \times 21$ target fixations on a grid $[-1, 1] \times [-1, 1]$ (normalized yaw/pitch coordinates), and generate trajectories to reach each of them. The final fixation point for each trajectory is shown as a distortion grid. The MSE in original coordinates is approximately $3.7$ degrees. (c) A subset of $5\times 5$ fixation points on the same grid (colored in red in the middle panel) is selected to show the individual trajectories generated to reach each target. Trajectories are divided into segments (alternating in red and blue color) to show the angular velocities in the yaw and pitch directions at each timestep. }
  \label{fig:combined_traj_generation}
\end{figure*}

\begin{figure*}[ht!]
  \centering
  \includegraphics[width=0.9\textwidth]{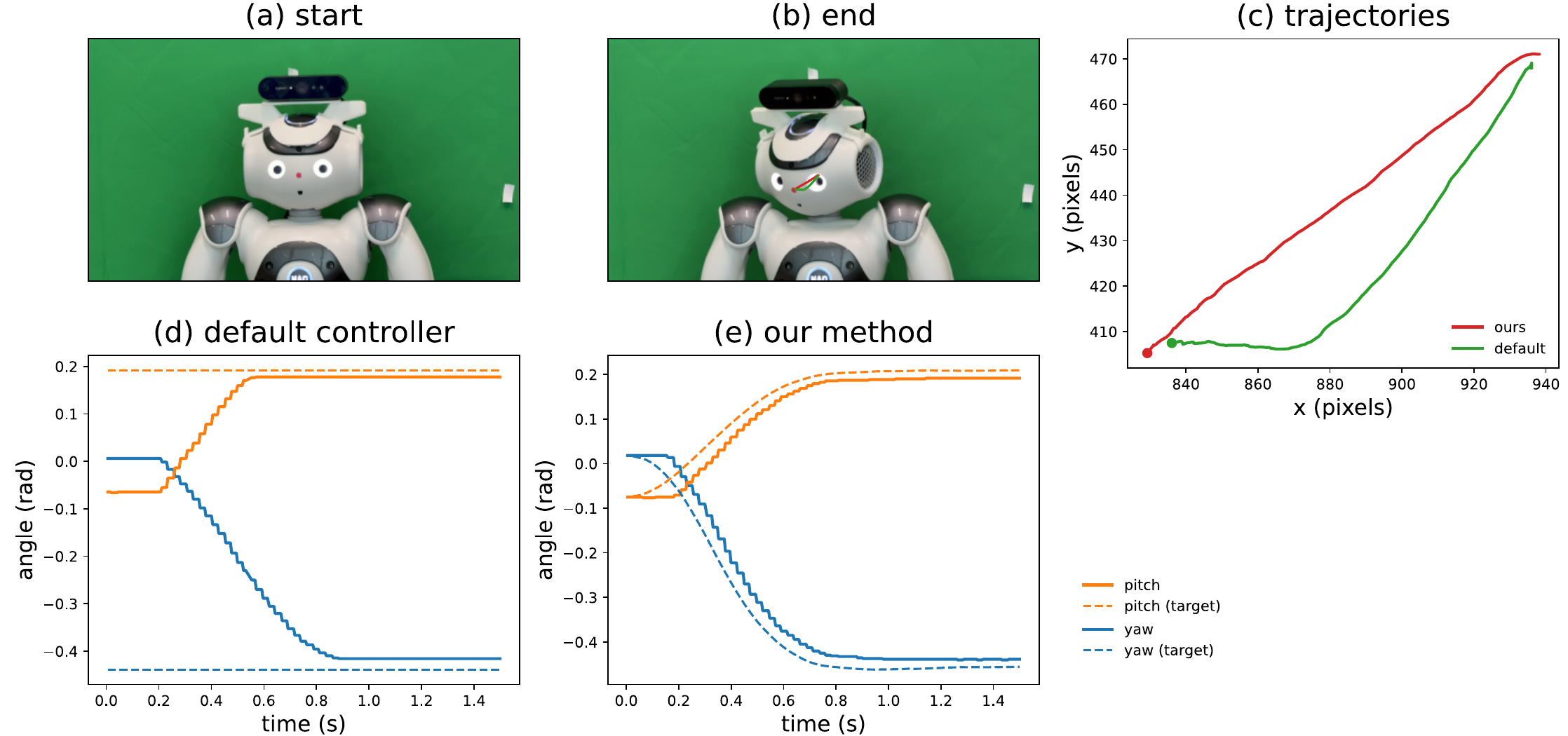}
  \caption{  Example of head movements on a Nao robot using our proposed method versus the default controller. (a-b) Initial and final pictures during a Nao fixation movement. (c) We track the red marker on Nao's nose (in pixel coordinates) during the fixation movement, using our method versus the default controller. (d-e) We report encoder readings from the Nao's head showing the robot's yaw and pitch angles while moving to a target configuration (d) or executing a trajectory (`target') generated using our proposed method. }
  \label{fig:nao_experiments1}
\end{figure*}

\section{Results} \label{results}

We evaluate the results of our approach in two stages. First, we look into the quality of the individual components to assess the quality of the generated head movement trajectories. We then evaluate the integrated system in a case study where a Nao robot needs to track a conversation between human co-participants.

\subsection{Trajectory modelling and generation}

We first performed a qualitative analysis of the VAE by generating 200 random trajectories (i.e., sampling random latent vectors to input into the trained decoder) and comparing them to the trajectories of human head movement from the dataset we collected (see Figure \ref{traj_all}). The generated trajectories are shown in Figure \ref{fig:combined_traj_generation} (a). A sample of individual trajectories reconstructed by the autoencoder is available in the Appendix as Figures \ref{traj_vae_reconstruction} and \ref{traj_vae_reconstruction_vel}.

We next evaluated whether trajectories to specific target fixations could be effectively generated using the trained MLP from Section \ref{sec:traj_generation}. We did so by selecting $21 \times 21$ (normalized) target fixation points within a uniform grid $[-1,1] \times [-1, 1]$, and generating trajectories to reach each of them. We then looked at the final fixation points reached by the predicted trajectories, and plotted them as a distortion grid in Figure \ref{fig:combined_traj_generation} (b). We found that the generated trajectories always reached the desired fixation point, with a mean square error of approximately $3.7^{\circ}$. We also found that the generated trajectories match the profile of angular velocities from the human data, as shown in Figure \ref{fig:combined_traj_generation} (c), which plots a sub-set of the trajectories and highlights the change in relative position at each timestep by alternating segments of different color. We also observed that even though the original dataset contained only few trajectories toward the four corners, our generative model managed to interpolate correctly and cover the whole range of possible fixation points accurately, suggesting that the model did not simply memorize the training trajectories, but rather managed to generalize well.

\subsection{Case study: active-speaker gazing task}

We evaluated the quality of the generated motions on a Nao robot during an active-speaker gazing task in two steps.

We first determined that the target trajectories generated using our method are effectively tracked by the robot motors, and that they are qualitatively different from a baseline where movements are obtained using the default Nao controller. We collected data in a group conversation setting by letting the robot passively participate in a conversation between three human interlocutors. During the conversations, we recorded videos centered on the robot's head together with the target trajectories generated with both methods and the joint angles as measured using the robot's encoders. We further extracted the movement of the head in the videos by tracking a red dot painted in the middle of the robot's face.

Figure \ref{fig:nao_experiments1} shows the results of this analysis. We found that the generated trajectories are significantly different from those generated by the baseline (panels (c), (d), and (e)), which we found to track the yaw and pitch movements independently of each other, so that in practice the pitch movement of the robot in the baseline terminates before the yaw movement, resulting in unnatural-looking motions.

\begin{figure}[ht!]
  \centering
  \includegraphics[width=0.3
\textwidth]{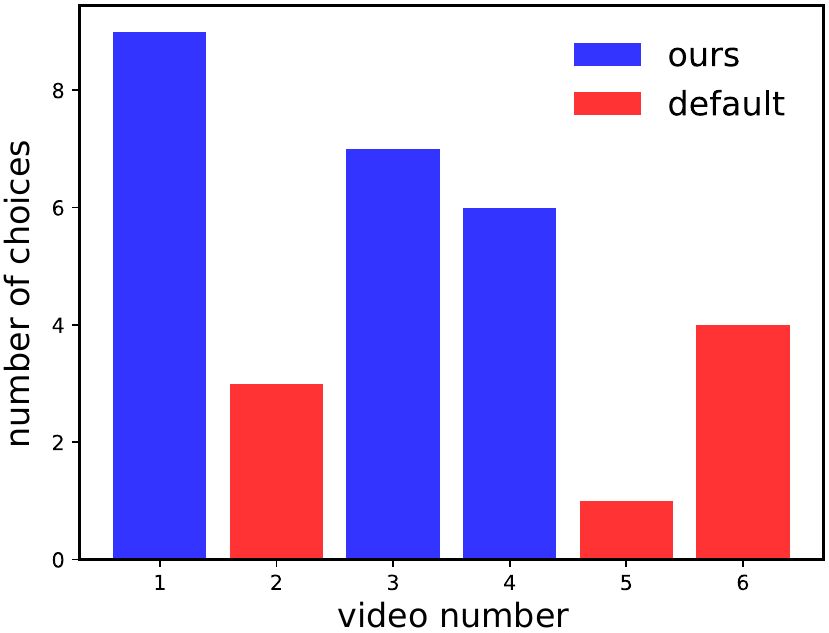}
  \caption{ Results from an experiment to assess human preferences for head motion generated using the proposed method versus a baseline default motion controller. Videos 1, 3, and 4 (blue bars) showed motion using our controller, while videos 2, 5, and 6 (red bars) were made using the baseline controller. Participants were asked to choose the 3 out of 6 videos that in their opinion showed the best looking and most natural movements. The histogram shows the total count of votes received for each video, suggesting that motion generated by our method is generally preferred ($6.33/9$) over the default motion controller ($2.67/9$).  }
  \label{fig:human_preferences}
\end{figure}

We further tested the subjective perception of the quality and naturalness of the generated motions by designing a preliminary experiment with $n=9$ human participants. Results are shown in Figure \ref{fig:human_preferences} as a histogram with the number of times each video was selected as within the top half of best-looking motions. Videos 1, 3, and 4 (blue) were generated using our method, and videos 2, 5, and 6 (red) used the default controller. We found that videos that used our method were chosen on average by $6.33$ out of 9 participants, while videos with the default controller were only chosen on average by $2.67$ participants.

\section{Discussion and Conclusions} \label{conclusion}

In this paper, we have addressed an often overlooked problem in human-robot interaction: imitating human head movements to track active speakers in a group conversation. Specifically, we addressed this problem by demonstrating how human data can be effectively used to generate head movements for a Nao robot by showing the efficacy of our motion controller in an active-speaker tracking task. The results indicate that the robot can move its head toward active speakers with natural-looking, well coordinated movements.

We suggest that this study could be further extended in the following ways. On the one hand, the same approach can be applied to upper body imitation for generating nonverbal cues (e.g., pointing, nodding, fidgeting) during the human-robot interaction, or to model more complex head movements with the inclusion of `roll' motion in addition to the current `yaw' and `pitch'. On the other hand, the method could be improved by training the whole system end-to-end to generate target trajectories directly from multimodal inputs, in an imitation learning setting.



\bibliographystyle{IEEEtran}
\bibliography{IEEEabrv,BosongAIRLab}

\begin{thebibliography}{10}
\providecommand{\url}[1]{#1}
\csname url@samestyle\endcsname
\providecommand{\newblock}{\relax}
\providecommand{\bibinfo}[2]{#2}
\providecommand{\BIBentrySTDinterwordspacing}{\spaceskip=0pt\relax}
\providecommand{\BIBentryALTinterwordstretchfactor}{4}
\providecommand{\BIBentryALTinterwordspacing}{\spaceskip=\fontdimen2\font plus
\BIBentryALTinterwordstretchfactor\fontdimen3\font minus
  \fontdimen4\font\relax}
\providecommand{\BIBforeignlanguage}[2]{{%
\expandafter\ifx\csname l@#1\endcsname\relax
\typeout{** WARNING: IEEEtran.bst: No hyphenation pattern has been}%
\typeout{** loaded for the language `#1'. Using the pattern for}%
\typeout{** the default language instead.}%
\else
\language=\csname l@#1\endcsname
\fi
#2}}
\providecommand{\BIBdecl}{\relax}
\BIBdecl

\bibitem{barot2023audio}
P.~Barot, E.~N. MacDonald, and K.~Mombaur, ``An audio-video sensor fusion
  framework to augment humanoid capabilities for identifying and interacting
  with human conversational partners,'' in \emph{2023 IEEE-RAS 22nd
  International Conference on Humanoid Robots (Humanoids)}.\hskip 1em plus
  0.5em minus 0.4em\relax IEEE, 2023, pp. 1--8.

\bibitem{physical_skills_from_videos}
X.~B. Peng, A.~Kanazawa, J.~Malik, P.~Abbeel, and S.~Levine, ``Sfv:
  Reinforcement learning of physical skills from videos,'' \emph{ACM
  Transactions On Graphics (TOG)}, vol.~37, no.~6, pp. 1--14, 2018.

\bibitem{npmp_humanoid}
J.~Merel, L.~Hasenclever, A.~Galashov, A.~Ahuja, V.~Pham, G.~Wayne, Y.~W. Teh,
  and N.~Heess, ``Neural probabilistic motor primitives for humanoid control,''
  \emph{arXiv preprint arXiv:1811.11711}, 2018.

\bibitem{laicago}
A.~Cully, J.~Clune, D.~Tarapore, and J.-B. Mouret, ``Robots that can adapt like
  animals,'' \emph{Nature}, vol. 521, no. 7553, pp. 503--507, 2015.

\bibitem{npmp_humanoid_and_quadruped}
S.~Bohez, S.~Tunyasuvunakool, P.~Brakel, F.~Sadeghi, L.~Hasenclever, Y.~Tassa,
  E.~Parisotto, J.~Humplik, T.~Haarnoja, R.~Hafner \emph{et~al.}, ``Imitate and
  repurpose: Learning reusable robot movement skills from human and animal
  behaviors,'' \emph{arXiv preprint arXiv:2203.17138}, 2022.

\bibitem{liu2022motor}
S.~Liu, G.~Lever, Z.~Wang, J.~Merel, S.~A. Eslami, D.~Hennes, W.~M. Czarnecki,
  Y.~Tassa, S.~Omidshafiei, A.~Abdolmaleki \emph{et~al.}, ``From motor control
  to team play in simulated humanoid football,'' \emph{Science Robotics},
  vol.~7, no.~69, p. eabo0235, 2022.

\bibitem{aloha2}
J.~Aldaco, T.~Armstrong, R.~Baruch, J.~Bingham, S.~Chan, K.~Draper, D.~Dwibedi,
  C.~Finn, P.~Florence, S.~Goodrich \emph{et~al.}, ``Aloha 2: An enhanced
  low-cost hardware for bimanual teleoperation,'' \emph{arXiv preprint
  arXiv:2405.02292}, 2024.

\bibitem{liao2023light}
J.~Liao, H.~Duan, K.~Feng, W.~Zhao, Y.~Yang, and L.~Chen, ``A light weight
  model for active speaker detection,'' in \emph{Proceedings of the IEEE/CVF
  Conference on Computer Vision and Pattern Recognition}, 2023, pp.
  22\,932--22\,941.

\bibitem{grassi2023robot}
L.~Grassi, C.~T. Recchiuto, and A.~Sgorbissa, ``Robot-induced group
  conversation dynamics: a model to balance participation and unify
  communities,'' in \emph{2023 IEEE/RSJ International Conference on Intelligent
  Robots and Systems (IROS)}.\hskip 1em plus 0.5em minus 0.4em\relax IEEE,
  2023, pp. 3991--3997.

\bibitem{chakravarty2016cross}
P.~Chakravarty and T.~Tuytelaars, ``Cross-modal supervision for learning active
  speaker detection in video,'' in \emph{Computer Vision ECCV 2016: 14th
  European Conference, Amsterdam, The Netherlands, October 11-14, 2016,
  Proceedings, Part V 14}.\hskip 1em plus 0.5em minus 0.4em\relax Springer,
  2016.

\bibitem{roth2020ava}
J.~Roth, S.~Chaudhuri, O.~Klejch, R.~Marvin, A.~Gallagher, L.~Kaver,
  S.~Ramaswamy, A.~Stopczynski, C.~Schmid, Z.~Xi \emph{et~al.}, ``Ava active
  speaker: An audio-visual dataset for active speaker detection,'' in
  \emph{ICASSP 2020-2020 IEEE International Conference on Acoustics, Speech and
  Signal Processing (ICASSP)}.\hskip 1em plus 0.5em minus 0.4em\relax IEEE,
  2020.

\bibitem{jung2024talknce}
C.~Jung, S.~Lee, K.~Nam, K.~Rho, Y.~J. Kim, Y.~Jang, and J.~S. Chung,
  ``Talknce: Improving active speaker detection with talk-aware contrastive
  learning,'' in \emph{ICASSP 2024-2024 IEEE International Conference on
  Acoustics, Speech and Signal Processing (ICASSP)}.\hskip 1em plus 0.5em minus
  0.4em\relax IEEE, 2024, pp. 8391--8395.

\bibitem{alcazar2022end}
J.~L. Alc{\'a}zar, M.~Cordes, C.~Zhao, and B.~Ghanem, ``End-to-end active
  speaker detection,'' in \emph{European Conference on Computer Vision}.\hskip
  1em plus 0.5em minus 0.4em\relax Springer, 2022, pp. 126--143.

\bibitem{gurvich2024real}
I.~Gurvich, I.~Leichter, D.~R. Palle, Y.~Asher, A.~Vinnikov, I.~Abramovski,
  V.~Gopal, R.~Cutler, and E.~Krupka, ``A real-time active speaker detection
  system integrating an audio-visual signal with a spatial querying
  mechanism,'' in \emph{ICASSP 2024-2024 IEEE International Conference on
  Acoustics, Speech and Signal Processing (ICASSP)}.\hskip 1em plus 0.5em minus
  0.4em\relax IEEE, 2024, pp. 8781--8785.

\bibitem{czarnecki2022we}
P.~Czarnecki and J.~Tkaczuk, ``As we speak: Real-time visually guided speaker
  separation and localization,'' in \emph{2022 IEEE 24th International Workshop
  on Multimedia Signal Processing (MMSP)}.\hskip 1em plus 0.5em minus
  0.4em\relax IEEE, 2022.

\bibitem{gilman1979measurement}
S.~Gilman, D.~D. Dirks, and S.~Hunt, ``Measurement of head movement during
  auditory localization,'' \emph{Behavior Research Methods \& Instrumentation},
  vol.~11, pp. 37--41, 1979.

\bibitem{lu2022sound}
H.~Lu and W.~O. Brimijoin, ``Sound source selection based on head movements in
  natural group conversation,'' \emph{Trends in Hearing}, vol.~26, p.
  23312165221097789, 2022.

\bibitem{vae}
D.~P. Kingma and M.~Welling, ``Auto-encoding variational bayes,'' in
  \emph{Proceedings of the International Conference on Learning
  Representations}, 2014.

\bibitem{mediapipe}
G.~LLC, ``Mediapipe: A framework for building perception pipelines,''
  \url{https://github.com/google/mediapipe}, 2019.

\end{thebibliography}

\appendix

\subsection{Supplementary Materials}
\label{appendix:suppl_materials}

We include the full set of hyperparameters used for training the variational autoencoder in Table \ref{tab:vae_hyperparams} (see Section \ref{sec:trajmodelling}), and the set of hyperparameters of the multilayer perceptron used to map target fixations into latent vectors in Table \ref{tab:fixation_mlp_hyperparams} (see Section \ref{sec:traj_generation}).

\begin{table}[ht!]
\centering
\caption{VAE Hyperparameters for Trajectory Modeling}
\label{tab:vae_hyperparams}
\begin{tabular}{lc}
\toprule
\textbf{Parameter}            & \textbf{Value} \\ 
\midrule
z-length                      & 10             \\ \addlinespace
Epochs                        & 500            \\ \addlinespace
Architecture                  & MLP with ReLU, 2 hidden layers [128, 128] \\ \addlinespace
Decoder Architecture          & MLP with ReLU, 2 hidden layers [128, 128] \\ \addlinespace
Batch Size                    & 16             \\ \addlinespace
Optimizer                     & Adam           \\ \addlinespace
Learning Rate                 & 0.001          \\ \addlinespace
Betas                         & (0.9, 0.999)   \\ 
\bottomrule
\end{tabular}
\end{table}

\begin{table}[ht!]
\centering
\caption{Hyperparameters for Fixation to Latent MLP in Trajectory Generation}
\label{tab:fixation_mlp_hyperparams}
\begin{tabular}{lc}
\toprule
\textbf{Parameter}            & \textbf{Value} \\ 
\midrule
Architecture                  & MLP, 2 hidden layers [64, 64], output size z-length=10 \\ \addlinespace
Epochs                        & 500            \\ \addlinespace
Optimizer                     & Adam           \\ \addlinespace
Learning Rate                 & 0.001          \\ \addlinespace
Batch Processing              & Full-batch     \\ 
\bottomrule
\end{tabular}
\end{table}

\subsection{Supplementary Results}

\begin{figure*}
  \centering
  \includegraphics[width=0.6\textwidth]{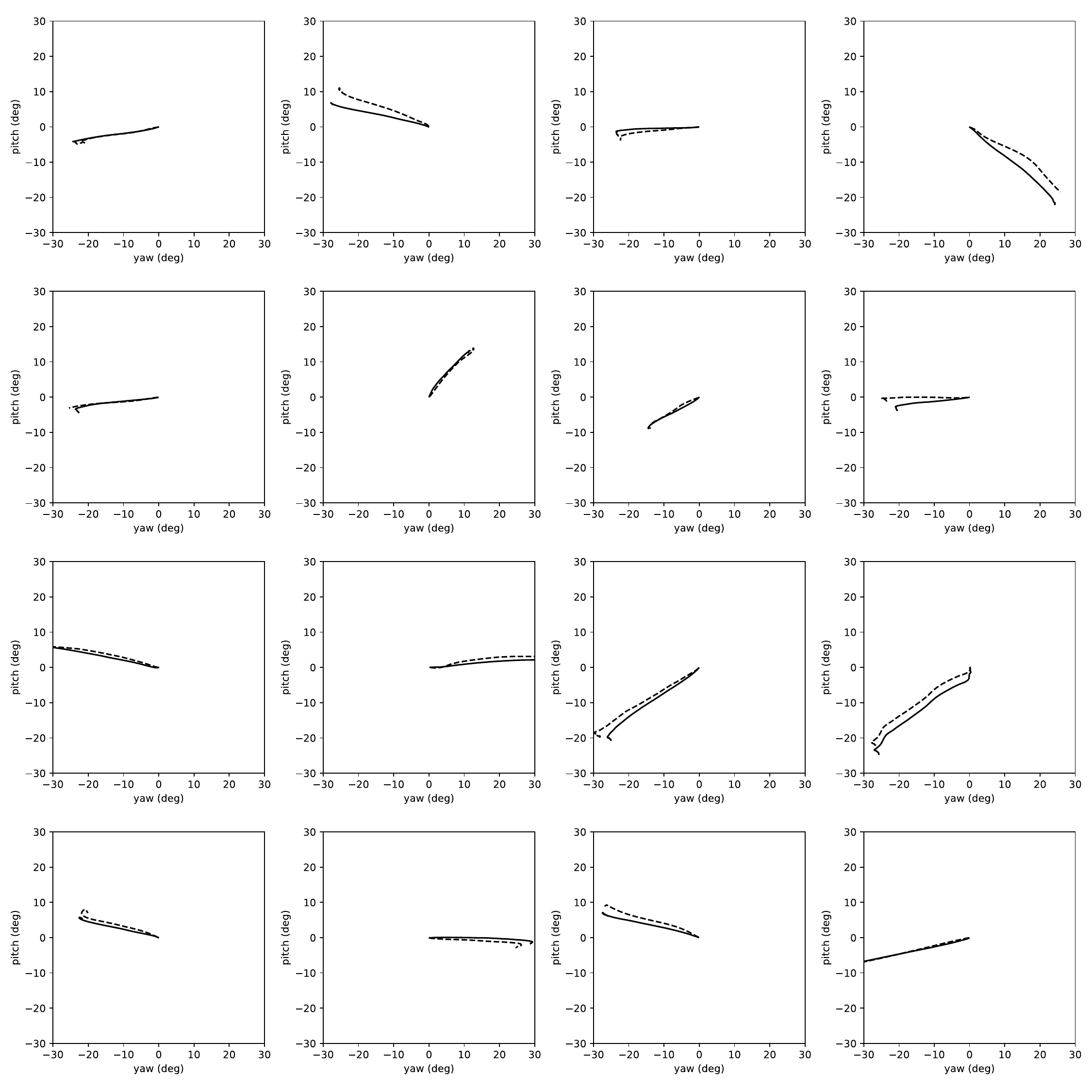}
  \caption{ Selection of 16 trajectories from the training set (dashed lines), together with their VAE-reconstruction (solid lines). }
  \label{traj_vae_reconstruction}
\end{figure*}

\begin{figure*}
  \centering
  \includegraphics[width=0.6\textwidth]{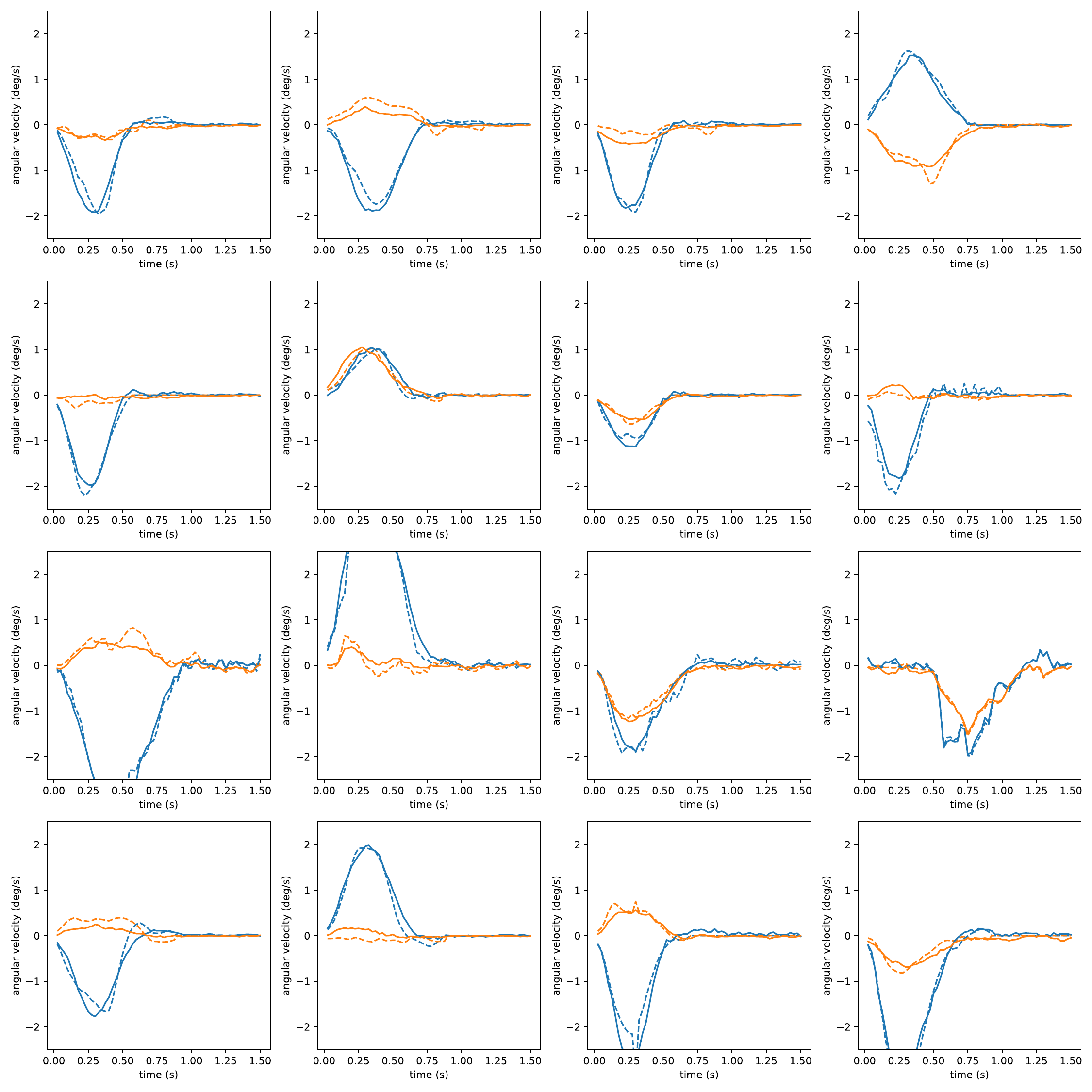}
  \caption{ Selection of 16 trajectories from the training set (dashed lines), together with their VAE-reconstruction (solid lines). Blue denotes yaw, and orange pitch. Trajectories are plotted as instantaneous velocities over time. }
  \label{traj_vae_reconstruction_vel}
\end{figure*}

We include a set of $16$ trajectories from the training set together with their corresponding reconstructed trajectories using the trained variational autoencoder. Results are shown in Figure \ref{traj_vae_reconstruction} as trajectories in (yaw, pitch) space, and in Figure \ref{traj_vae_reconstruction_vel} as instantaneous angular velocities.

\end{document}